%% file: main.tex
\documentclass[runningheads]{llncs}
\usepackage{graphicx}
\usepackage{amsmath,amssymb} 
\usepackage{color}
\usepackage{hyperref}

\usepackage{graphicx}
\usepackage{amsmath,amssymb} 
\usepackage{color}
\usepackage{booktabs}
\usepackage{gensymb}
\usepackage{multirow}
\usepackage{subfig}
\usepackage{makecell}

\def\blfootnote{\xdef\@thefnmark{}\@footnotetext}

\begin{document}
\title{Deep Model-Based 6D Pose Refinement in RGB} 

\titlerunning{Deep Model-Based 6D Pose Refinement in RGB}
%
\author{Fabian Manhardt\inst{1}$^{*}$ \and Wadim Kehl\inst{2}$^{*}$ \and Nassir Navab\inst{1} \and Federico Tombari\inst{1}}

%
\authorrunning{F. Manhardt, W. Kehl, N. Navab and F. Tombari}
%

\institute{Technical University of Munich, Garching b. Muenchen 85748, Germany \\ \email{\{fabian.manhardt, nassir.navab\}@tum.de} \quad \email{tombari@in.tum.de}
\and
Toyota Research Institute, Los Altos, CA 94022, USA\\
\email{wadim.kehl@tri.global}}
%

\maketitle

\begin{abstract}
	We present a novel approach for model-based 6D pose refinement in color data. Building on the established idea of contour-based pose tracking, we teach a deep neural network to predict a translational and rotational update. At the core, we propose a new visual loss that drives the pose update by aligning object contours, thus avoiding the definition of any explicit appearance model. In contrast to previous work our method is correspondence-free, segmentation-free, can handle occlusion and is agnostic to geometrical symmetry as well as visual ambiguities. Additionally, we observe a strong robustness towards rough initialization. The approach can run in real-time and produces pose accuracies that come close to 3D ICP without the need for depth data. Furthermore, our networks are trained from purely synthetic data and will be published together with the refinement code at \url{http://campar.in.tum.de/Main/FabianManhardt} to ensure reproducibility. 
	\keywords{Pose Estimation, Pose Refinement, Tracking}
\end{abstract}
{\let\thefootnote\relax\footnotetext{* The first two authors contributed equally to this work.}}
\input{intro}	
\input{related}	
\input{method}

\input{eval}

\section{Conclusion}
We believe to have presented a new approach towards 6D model tracking in RGB with the help of deep learning and we demonstrated the power of our approach on multiple datasets and for the scenarios of pose refinement and for instance/category tracking. Future work will include investigation towards generalization to other domains, e.g. the suitability towards visual odometry.\newline

\noindent  \textbf{Acknowledgments}
We would like to thank Toyota Motor Corporation for funding and supporting this work.

\clearpage

\bibliographystyle{splncs04}
\bibliography{egbib}

\end{document}

%% file: intro.tex
\section{Introduction}
\begin{figure}[t!]
	\begin{tabular}{ccc}
		\includegraphics[width=3.95cm]{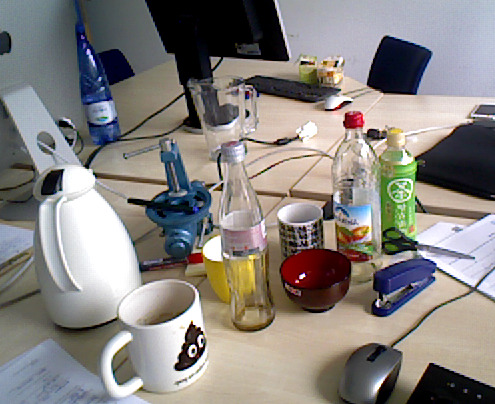} &
		\includegraphics[width=3.95cm]{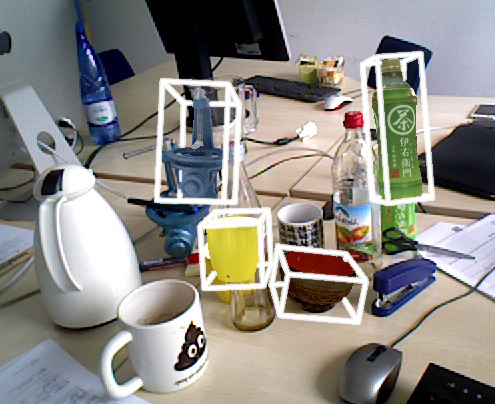} & 
		\includegraphics[width=3.95cm]{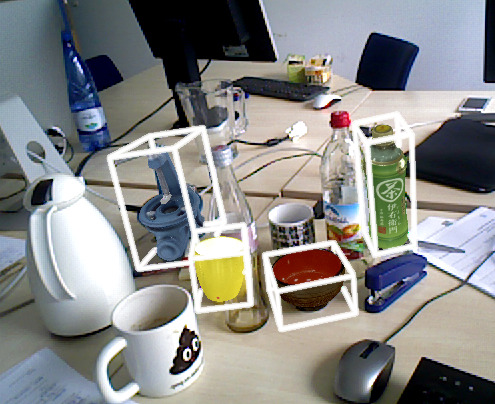} \\
		a) Input Image & b) Initial pose hypotheses & c) Poses after 10 iterations
	\end{tabular}
	\caption{Exemplary illustration of our method. While a) depicts an input RGB frame, b) shows our four initial 6D pose hypotheses. For each obtained frame we refine each pose for a better fit to the scene. In d) we show the final results after convergence. Note the rough pose initializations as well as the varying amount of occlusion the objects of interest undergo. }
	\label{fig:teaser}
\end{figure}

The problem of tracking CAD models in images is frequently encountered in contexts such as robotics, augmented reality (AR) and medical procedures.  Usually, tracking has to be carried out in the full 6D pose, i.e. one seeks to retrieve both the 3D metric translation as well as the 3D rotation of the object in each frame. Another typical scenario is pose refinement, where an object detector provides a rough 6D pose estimate, which has to be corrected in order to provide a better fit (Figure \ref{fig:teaser}). The usual difficulties that arise include viewpoint ambiguities, occlusions, illumination changes and differences in appearance between the model and the object in the scene. Furthermore, for tracking applications the method should also be fast enough to cover large inter-frame motions. 

Most related work based on RGB data can be roughly divided into sparse and region-based methods. The former methods try to establish local correspondences between frames \cite{Vacchetti2004,Park2008} and work well for textured objects, whereas latter ones exploit more holistic information about the object such as shape, contour or color \cite{Prisacariu2012,Dambreville2010,Tjaden2016,Tjaden2017} and are usually better suited for texture-less objects. It is worth mentioning that mixtures of the two sets of methods have been proposed as well \cite{Schmaltz2007,Brox2010,Schmaltz2012,Pauwels2013}. Recently, methods that use only depth \cite{Tan2015} or both modalities \cite{Krull2014,Kehl2017a,Garon2017} have shown that depth can make tracking more robust by providing more clues about occlusion and scale. 

This work aims to explore how RGB information alone can be sufficient to perform visual tasks such as 3D tracking and 6-Degree-of-Freedom (6DoF) pose refinement by means of a Convolutional Neural Network (CNN). While this has already been proposed for camera pose and motion estimation \cite{Kendall2017,Zhou2017,Wang2017,Ummenhofer2017}, it has not been well-studied for the problem at hand.
 
As a major contribution we provide a differentiable formulation of a new visual loss that aligns object contours and implicitly optimizes for metric translation and rotation. While our optimization is inspired by region-based approaches, we can track objects of any texture or shape since we do not need to model global \cite{Prisacariu2012,Tjaden2016,Kehl2017a} or local appearance \cite{Hexner2016,Tjaden2017}. Instead, we show that we can do away with these hand-crafted approaches by letting the network learn the object appearance implicitly.  We teach the CNN to align contours between synthetic object renderings and scene images under changing illumination and occlusions and show that our approach can deal with a variety of shapes and textures. Additionally, our method allows to deal with geometrical symmetries and visual ambiguities without manual tweaking and is able to recover correct poses from very rough initializations.  

Notably, our formulation is parameter-free and avoids typical pitfalls of hand-crafted tracking or refinement methods (e.g. via segmentation or correspondences + RANSAC) that require tedious tuning to work well in practice. Furthermore, like with depth-based approaches such as ICP, we are robust to occlusion and produce results which come close to RGB-D methods without the need for depth data, making it thus very applicable to the domains of AR, medical and robotics.

%% file: related.tex
\section{Related work}
Since the field of tracking and pose refinement is vast, we will only focus here on works that deal with CAD models in RGB data.
Early methods in this field used either 2D-3D correspondences \cite{Rosenhahn2006,Schmaltz2007} or 3D edges \cite{Drummond2002,Tateno2009,Seo2014} and fit the model in an ICP fashion with iterative, projective update steps. Successive methods in this direction managed to obtain improved performance \cite{Brox2010,Schmaltz2012}. Additionally, other works focused on tracking the contour densely via level-sets \cite{Bibby2008,Dambreville2010}. 

Based on these works, \cite{Prisacariu2012} presented a new approach that follows the projected model contours to estimate the 6D pose update. In a follow-up work \cite{Prisacariu2015}, the authors extended their method to simultaneously track and reconstruct a 3D object on a mobile phone in real-time. The authors from \cite{Tjaden2016} improved the convergence behavior with a new optimization scheme and presented a real-time implementation on a GPU. Consequently, \cite{Tjaden2017} showed how to improve the color segmentation by using local color histograms over time. Orthogonally, the work \cite{Kehl2017a} approximates the model pose space to avoid GPU computations and enables real-time performance on a single CPU core. All these approaches share the property that they rely on hand-crafted segmentation methods that can fail in the case of sudden appearance changes or occlusion. We instead want to entirely avoid hand-crafting manual appearance descriptions. 

Another set of works tries to combine learning with simultaneous detection and pose estimation in RGB. The method presented in \cite{Kehl2017} couples the SSD paradigm \cite{Liu2016} with pose estimation to produce 6D pose pools per instance which are then refined with edge-based ICP. On the contrary, the approach from \cite{Brachmann2016} uses auto-context Random Forests to regress object coordinates in the scene that are used to estimate poses. In \cite{Rad2017} a method is presented that instead regresses the projected 3D bounding box and recovers the pose from these 2D-3D correspondences whereas the authors in \cite{Pavlakos2017} infer keypoint heatmaps that are then used for 6D pose computation. Similarly, the 3D Interpreter Network \cite{Wu2016} infers heatmaps for categories and regresses projection and deformation to align synthetic with real imagery. In the work \cite{Garon2017}, a deep learning approach is used to track models in RGB-D data. Their work goes along similar grounds but we differ in multiple ways including data generation, energy formulation and their use of RGB-D data. In particular, we show that a naive formulation of pose regression does not work in the case of symmetry which is often the case for man-made objects. 

We also find common ground with Spatial Transformer Networks in 2D \cite{Jaderberg2015} and especially 3D \cite{Bhagavatula2017}, where the employed network architecture contains a submodule to transform the 2D/3D input via a regressed affine transformation on a discrete lattice. Our network instead regresses a rigid body motion on a set of continuous 3D points to minimize the visual error.

%% file: method.tex
\section{Methodology}

\begin{figure}[t!]
	\centering
	\includegraphics[width=12cm]{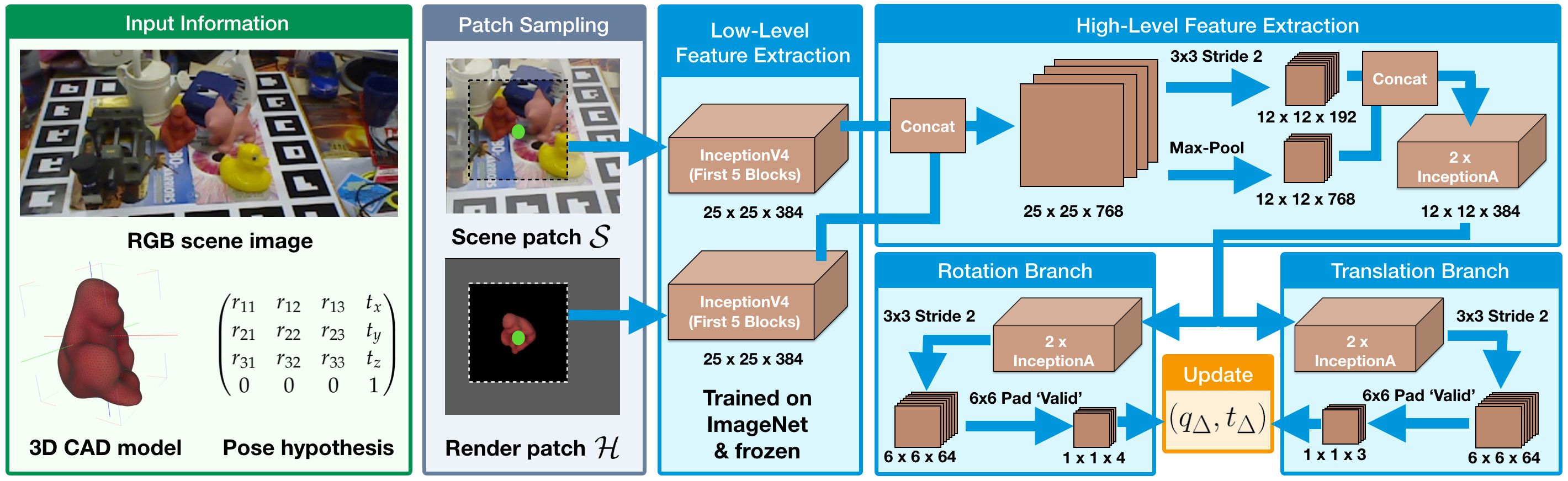}
	\caption{Schematic overview of the full pipeline. Given input image and pose hypothesis $(R, t)$, we render the object, compute the center of the bounding box of the hypothesis (green point) and then cut out a scene patch $\mathcal{S}$ and a render patch $\mathcal{H}$. We resize both to 224x224 and feed them separately into pre-trained InceptionV4 layers to extract low-level features. Thereafter, we concatenate and compute high-level features before diverging into separate branches. Eventually, we retrieve our pose update as 3D translation and normalized 4D quaternion.}	
	\label{fig:network}
\end{figure}

In this section we explain our approach to train a CNN to regress a 6D pose refinement from RGB information alone. We design the problem in such a way that we supply two color patches ($\mathcal{S}$ and $\mathcal{H}$) to the network in order to infer a translational and rotational update. In Figure \ref{fig:network} we depict our pipeline and show a typical scenario where we have a 6D hypothesis (coming from a detector or tracker) that is not correctly aligned. We want to estimate a refinement such that eventually the updated hypothesis overlaps perfectly with the real object.

\subsection{Input patch sampling}
We first want to discuss our patch extraction strategy. Provided a CAD model and a 6D pose estimate $(R,t)$ in camera space, we create a rendering and compute the center of the associated bounding box of the hypothesis around which we subsequently extract $\mathcal{S}$ and $\mathcal{H}$. Since different objects have varying sizes and shapes it is important to adapt the cropping size to the spatial properties of the specific object. The most straightforward method would be to simply crop $\mathcal{S}$ and $\mathcal{H}$ with respect to a tight 2D bounding box of the rendered mask. However, when employing such metric crops, the network loses the ability to robustly predict an update along the Z-axis: indeed, since each crop would almost entirely fill out the input patch, no estimate of the difference in depth can be drawn. Due to this, we explicitly calculate the spatial extent in pixels at a minimum metric distance (with some added padding) and use this as a fixed-size 'window' into our scene. In particular, prior to training, we render the object from various different viewpoints, compute their bounding boxes, and take the maximum width or height of all produced bounding boxes.

\subsection{Training stage}

To create training data we randomly sample a ground truth pose $(R^*, t^*)$ of the object in camera coordinates and render the object with that pose onto a random background to create a scene image. To learn pose refinement, we perturb the true pose to get a noisy version $(R,t)$ and render a hypothesis image. Given those two images, we cut out patches $\mathcal{S}$ and $\mathcal{H}$ with the strategy mentioned above.

\subsubsection{The naive approach}

Provided these patches, we now want to infer a separate correction $(R_\Delta, t_\Delta)$ of the perturbed pose $(R,t)$ such that 
\begin{equation}
R^* = R_\Delta \cdot R, \quad t^* = t + t_\Delta.
\end{equation}
Due to the difficulty of optimizing in SO(3) we parametrize via unit quaternions $q^*, q, q_\Delta$ to define a regression problem, i.e. similar to what \cite{Kendall2015} proposed for camera localization or \cite{Garon2017} for model pose tracking:
\begin{equation}
\min_{q_\Delta,t_\Delta} \big| \big| q^* - \frac{q_\Delta}{||q_\Delta||} \big| \big| + \gamma \cdot \big| \big| t^* - t_\Delta \big| \big|
\end{equation}
In essence, this energy weighs the numerical error in rotation against the one in translation by means of the hyper-parameter $\gamma$ and can be optimized correctly when solutions are unique (as is the case, e.g., of camera pose regression). Unfortunately, the above formulation only works for injective relations where an input image pair gets always mapped to the same transformation. In the case of one-to-many mappings, i.e. an image pair can have multiple correct solutions, the optimization does not converge since it is pulled into multiple directions and regresses the average instead. In the context of our task, visual ambiguity is common for most man-made objects because they are either symmetric or share the same appearance from multiple viewpoints. For these objects there is a large (sometimes infinite) set of refinement solutions that yield the same visual result. In order to regress $q_\Delta$ and $t_\Delta$ under ambiguity, we therefore propose an alternative formulation. 
 
\subsubsection{Proxy loss for visual alignment}

Instead of explicitly minimizing an ambiguous error in transformation, we strive to minimize an unambiguous error that measures similarity in appearance. We thus treat our search for the pose refinement parameters as a subproblem inside another proxy loss that optimizes for visual alignment. While there are multiple ways to define a similarity measure, we seek one that fulfills the following properties: 1) invariant to symmetric or indistinguishable object views, 2) robust to color deviation, illumination change and occlusion as well as 3) smooth and differentiable with respect to the pose.

\begin{figure}[t!]
	\centering
	\begin{tabular}{cccc}
		\includegraphics[width=2.9cm]{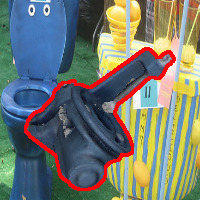} &
		\includegraphics[width=2.9cm]{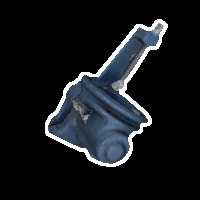} & 
		\includegraphics[width=2.9cm]{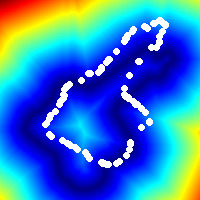} &
		 \includegraphics[width=2.9cm]{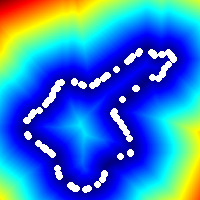}
		 \\
		\makecell{a) Synthetic scene \\ input image $\mathcal{S}$}  & \makecell{b) 6D hypothesis  \\ rendering $\mathcal{H}$} & \makecell{c) Pose estimate at  \\ initial training state} & \makecell{d) Refinement after \\ convergence}
	\end{tabular}
	
	\caption{Visualization of our training procedure. In (a) and (b) we show the two image patches that constitute one training sample and the input to our network. We highlight for the reader the contours for which we seek the projective alignment from white to red. In (c) we see the initial state of training with no refinement together with the distance transform of the scene $\mathcal{D}_\mathcal{S}$ and the projection of 3D sample points $V_\mathcal{H}$ from the initial 6D hypothesis. Finally, in (d) we can see the refinement after convergence. }
	\label{fig:refinement}
\end{figure}

To fulfill the first two properties we propose to align the object contours. Tracking the 6D pose of objects via projective contours has been presented before \cite{Kehl2017a,Tjaden2016,Prisacariu2012} but, to the best of our knowledge, has not so far been introduced in a deep learning framework. Contour tracking allows to reduce the difficult problem of 3D geometric alignment to a simpler task of 2D silhouette matching by moving through a distance transform, avoiding explicit correspondence search. Furthermore, a physical contour is not affected by deviations in coloring or lighting which makes it even more appealing for pure RGB methods. We refer to Figure \ref{fig:refinement} for a training example and the visualization of the contours we align.

Fulfilling smoothness and differentiability is more difficult. An optimization step for this energy requires to render the object with the current pose hypothesis for contour extraction, estimate the similarity with the target contour and back-propagate the error gradient such that the refined hypothesis' projected contour is closer in the next iteration. Unfortunately, back-propagating through a rendering pipeline is non-trivial (due to, among others, z-buffering and rasterization). We therefore propose here a novel formulation to drive the network optimization successfully through the ambiguous 6D solution space. We employ an idea, introduced in \cite{Kehl2017a}, that allows us to use an approximate contour for optimization without iterative rendering. When creating a training sample, we use the depth map of the rendering to compute a 3D point cloud in camera space and sample a sparse point set on the contour, denoted as $V := \{v \in \mathbb{R}^3 \}$. The idea is then to transform these contour points with the current refinement estimate $(q_\Delta,t_\Delta)$, followed by a projection into the scene. This mimics a rendering plus contour extraction at no cost and allows for back-propagation. 

For a given training sample with input patch pair $(\mathcal{S}, \mathcal{H})$, a distance transform of the scene contour $\mathcal{D}_\mathcal{S}$ and hypothesis contour points $V_\mathcal{H}$, we define the loss
\begin{equation}
\mathcal{L}(q_\Delta, t_\Delta, \mathcal{D}_\mathcal{S}, V_\mathcal{H}) := \sum_{v \in V_\mathcal{H}} \mathcal{D}_\mathcal{S} \bigg [\pi \big( q_\Delta \cdot v \cdot q^{-1}_\Delta + t_\Delta \big) \bigg ]
\end{equation}
with $q^{-1}_\Delta$ being the conjugate quaternion. With the formulation above we also free ourselves from any $\gamma$-balancing issue between quaternion and translation magnitudes as in a standard regression formulation.

Minimizing the above loss with a gradient descent step forces a step towards the 0-level set of the distance transform. We basically tune the network weights to rotate and translate the object in 6D to maximize the projected contour overlap. While this works well in practice, we have observed that for certain objects and stronger pose perturbations the optimization can get stuck in local minima. This occurs when our loss drives the contour points into a configuration where the distance transform allows them to settle in local valleys. To remedy this problem we introduce a bi-directional loss formulation that simultaneously aligns the contours of hypothesis as well as scene onto each other, coupled and constrained by the same pose update. We thus have an additional term that runs into the opposite direction:
\begin{equation}
\mathcal{L} := \mathcal{L}(q_\Delta, t_\Delta, \mathcal{D}_\mathcal{S}, V_\mathcal{H}) + \mathcal{L}(q^{-1}_\Delta, -t_\Delta, \mathcal{D}_\mathcal{H}, V_\mathcal{S}).
\end{equation}
This final loss $\mathcal{L}$ does not only alleviate the locality problem but has also shown to lead to faster training overall. We therefore chose this energy for all experiments.

\subsection{Network design and implementation}
We give a schematic overview of our network structure in Figure \ref{fig:network} and provide here more details. In order to ensure fast inference, our network follows a fully-convolutional design. The network is fed with two $224\times224\times3$ input patches representing the cropped scene image $\mathcal{S}$ and cropped render image $\mathcal{H}$. Both patches run in separate paths through the first levels of an InceptionV4 \cite{Szegedy2016} instance to extract low-level features. Thereafter we concatenate the two feature tensors, down-sample by employing max-pooling as well as a strided $3\times3$ convolution, and concatenate the results again. After two Inception-A blocks we branch off into two separate paths for the regression of rotation and translation. In each we employ two more Inception-A blocks before down-sampling by another strided $3\times3$ convolution. The resulting tensors are then convolved with either a $6\times6\times4$ kernel to regress a 4D quaternion or a $6\times6\times3$ kernel to predict a 3D update translation vector.

Initial experiments showed clearly that training the network from scratch made it impossible to bridge the domain gap between synthetic and real images. Similarly to \cite{Kehl2017,Hinterstoisser2017} we found that the network focused on specific appearance details of the rendered CAD models and the performance on real imagery collapsed drastically. Synthetic images usually possess very sharp edges and clear corners. Since the first layers learn low-level features they overfit quickly to this perfect rendered world during training. We therefore copied the first five convolutional blocks from a pre-trained model
and froze their parameters. We show the improvements in terms of generalization to real data in the supplement.


Further, we initialize the final regression layers such that the bias equals identity quaternion and zero translation whereas the weights are given a small Gaussian noise level of $\sigma=0.001$. This ensures that we start refinement from a neutral pose, which is crucial for the evaluation of the projective visual loss. 

While our approach produces very good refinements in a single shot we decided to also implement an iterative version where we run the pose refinement multiple times until the regressed update falls under a threshold. 

%% file: eval.tex
\section{Evaluation}

We ran our method with TensorFlow 1.4 \cite{Abadi2016} on a i7-5820K@3.3GHz with an NVIDIA GTX 1080. For all experiments we ran the training with 100k iterations, a batch size of 16 and ADAM with a learning rate of $3 \cdot 10^{-4}$. Furthermore, we fixed the number of 3D contour points per view to $|V_\mathcal{S}| = |V_\mathcal{H}| = 100$. Additionally, our method is real-time capable since one iteration requires approximately 25ms during testing.

To evaluate our method, we carried out experiments on three, both synthetic and real, datasets and will convey that our method can come close to RGB-D based approaches. In particular, the first dataset, referred to as 'Hinterstoisser', was introduced in \cite{Hinterstoisser2012} and consists of 15 sequences each possessing approximately 1000 images with clutter and mild occlusion. Only 13 of these provide water-tight CAD models and we therefore, like others before us, skip the other two sequences. The second one, which we refer to as 'Tejani', was proposed in \cite{Tejani2014} and consists of six mostly semi-symmetric, textured objects each undergoing different levels of occlusion. In contrast to the first two real datasets, the latter one, referred to as 'Choi' \cite{Choi2013}, consists of four synthetic tracking sequences.

In essence, we will first conduct some self-evaluation in which we illustrate our convergence properties with respect to different degrees of pose perturbation on real data. Then we show our method when applied to object tracking on 'Choi'. As a second application, we compare our approach to a variety of other state-of-the-art RGB and RGB-D methods by conducting experiments in pose refinement on 'Hinterstoisser', the 'Occlusion' dataset and 'Tejani'. Finally, we depict some failure cases and conclude with a qualitative category-level experiment.

\subsection{Pose perturbation}

\begin{figure}[t!]
	\captionsetup[subfigure]{labelformat=empty}
	\centering
		\includegraphics[width=10cm]{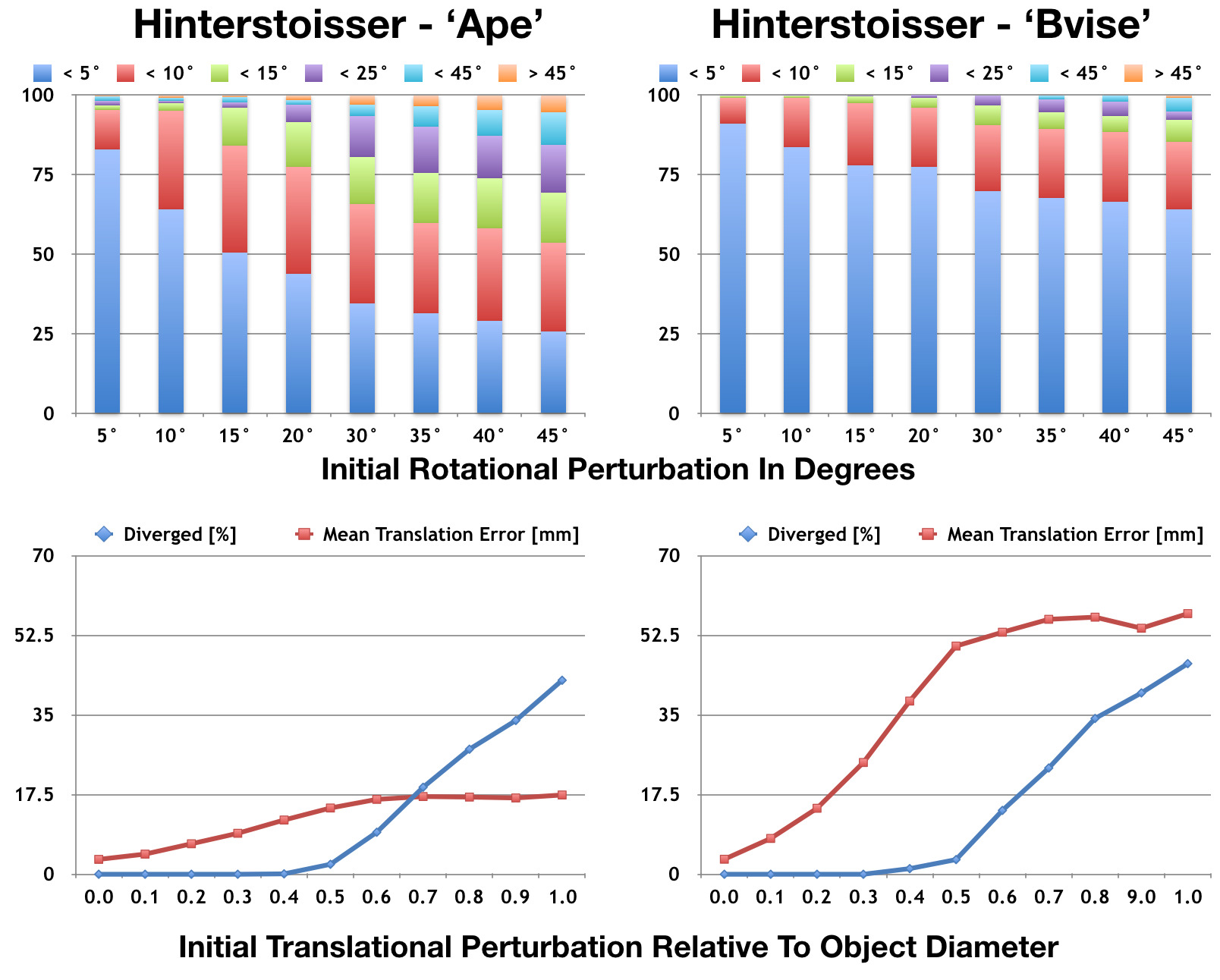}
		\setcounter{subfigure}{0}
		\subfloat[Perturbation]{\includegraphics[width=2.9cm]{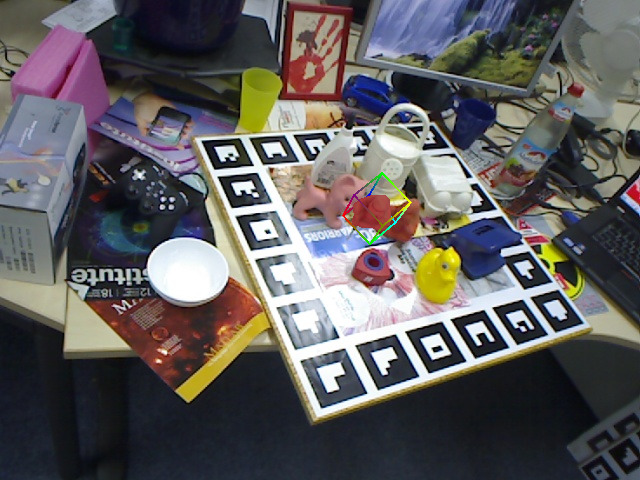}}  
		\subfloat[Refinement]{\includegraphics[width=2.9cm]{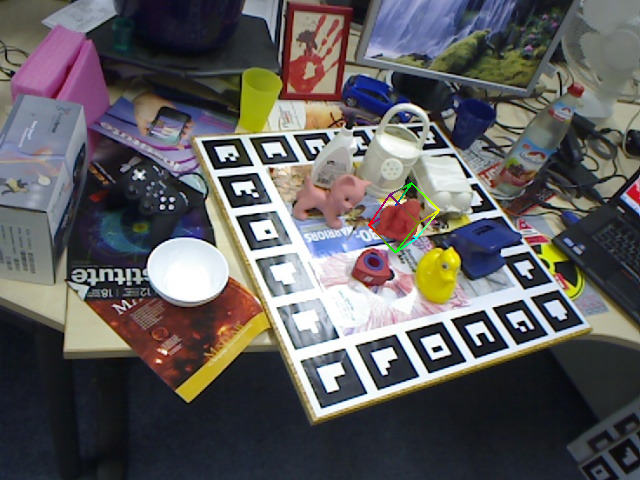}}
		\setcounter{subfigure}{0}
		\subfloat[Perturbation]{\includegraphics[width=2.9cm]{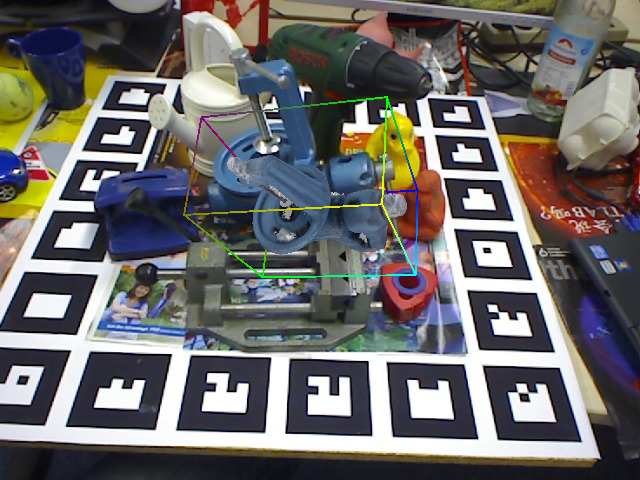}}  
		\subfloat[Refinement]{\includegraphics[width=2.9cm]{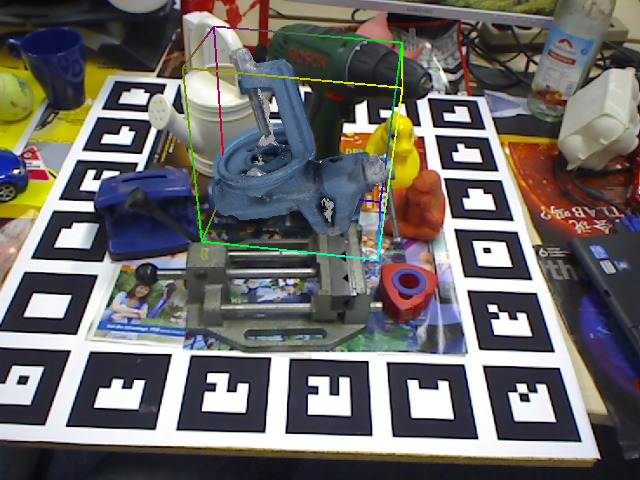}} \\

	\caption{Top: Perturbation results for two objects from \cite{Hinterstoisser2012} for increasing rotation and translation levels. Bottom: Qualitative results from the same experiment.}
	\label{fig:pertubations}
\end{figure}
We study the convergence behavior of our method by taking correct poses, applying a perturbation by a certain amount and measure how well we can refine back to the original pose. To this end, we use the 'Hinterstoisser' dataset since it provides a lot of variety in terms of both colors and shapes. For each frame of a particular sequence we perturb the ground truth pose either by an angle or by a translation vector. In Figure \ref{fig:pertubations} we illustrate our results for the 'ape' and the 'bvise' objects and kindly refer the reader to the supplement for all graphs. In particular, we report our results for increasing degrees of angular perturbations from 5\degree to 45\degree and for increasing translation perturbations from 0 to 1 relative to the object's diameter. We define divergence if the refined rotation is above 45\degree in error or the refined translation larger than half of the object's diameter and we employ 10 iterative steps to maximize our possible precision. 

In general, our method can recover poses very robustly even under strong perturbations. Even for the extreme case of rotating the 'bvise' with 45\degree we can refine back to an error less than 5\degree in more than 60\% of all trials, and to an error less than 10\degree in more than 80\% of all runs. Additionally, our approach only diverged for less than 1\%. However, for the more difficult 'ape' object our numbers worsen. In particular, in almost 50\% of the cases we were not able to rotate back the object to an error of less than 10\%. Yet, this can be easily explained by the object's appearance. The 'ape' is a rather small object with poor texture and non-distinctive shape, which does not provide enough information to hook onto whereas the 'bvise' is large and rich in appearance. It is noteworthy that the actual divergence behavior in rotation is similar for both and that the visual alignment for the 'ape' is often very good despite the error in pose. 

The translation error correlates almost linearly between initial and final pose. We also observe an interesting tendency starting from perturbation levels at around 0.6 after which the results can be divided up into two distinct sets: either the pose diverges or the error settles on a certain level. This implies that certain viewpoints are easy to align as long as they have a certain visual overlap to begin with, rather independent of how strong we perturb. Other views instead are more difficult with higher perturbations and diverge from some point on.

\subsection{Tracking}

\begin{figure}[t!]
	\centering
	\begin{tabular}{cc}
		\scalebox{0.75}{
			\begin{tabular}[b]{c|c|c|c|c|c|c|c|c}
				
				\multicolumn{1}{c}{} &
				\multicolumn{1}{c}{} &
				\multicolumn{1}{c}{PCL} &
				\multicolumn{1}{c}{C\&C} &
				\multicolumn{1}{c}{Krull} &
				\multicolumn{1}{c}{Tan} &
				\multicolumn{1}{c}{Kehl} &
				\multicolumn{1}{c}{Tjaden} &
				\multicolumn{1}{c}{Ours} 
				\\
				\midrule
				\multirow{6}{*}{{\rotatebox[origin=c]{90}{(a)~\emph{Kinect Box}}}}
				&$t_x$    &    43.99    &    1.84    & 0.8    &   1.54   &    \textbf{0.76} & 55.75 & 1.46 \\
				&$t_y$    &    42.51    &    2.23    & 1.67   &   1.90  &  \textbf{1.09} & 70.57 & 2.28 \\
				&$t_z$    &    55.89    &    1.36    & 0.79   &   \textbf{0.34}  & 0.38 &  402.14 & 10.61\\  			
				&\emph{$\alpha$}&    7.62 &    6.41  &  1.11  &    0.42  & \textbf{0.17} & 42.61 & 1.84  \\      		
				&\emph{$\beta$} &    1.87    &    0.76 &        0.55        &  0.22   &  \textbf{0.18} &  27.74 & 2.09 \\
				&\emph{$\gamma$}&    8.31    &    6.32    &     1.04        &    0.68    &  \textbf{0.20} & 38.979 & 1.23 \\
				\midrule
				\multirow{6}{*}{{\rotatebox[origin=c]{90}{(b)~\emph{Milk}}}}
				&$t_x$    &    13.38    &    0.93    & \textbf{    0.51    }    &        1.23   & 0.64 &  39.21 & 3.89 \\
				&$t_y$    &    31.45    &    1.94    & 1.27        &        0.74     &      \textbf{0.59} & 48.13 & 4.25 \\
				&$t_z$    &    26.09    &    1.09    & 0.62        &    \textbf{    0.24    }    &   \textbf{0.24    } &  332.11 &57.68 \\
				&\emph{$\alpha$}    &    59.37    &    3.83 &        2.19        &    0.50  &  \textbf{0.41} &  45.54 &  38.74\\ 
				&\emph{$\beta$}    &    19.58    &    1.41 &        1.44        &    \textbf{0.28    } &   0.29 & 26.37 & 27.62 \\
				&\emph{$\gamma$}&    75.03    &    3.26    &     1.90        &   0.46      &  \textbf{0.42} & 21.72 & 42.68 \\
				\midrule
				\multirow{6}{*}{{\rotatebox[origin=c]{90}{(c)~\emph{Orange Juice}}}}
				&    $t_x$    &    2.53    &    0.96    &0.52        &        1.10        &    \textbf{0.50} & 2.29 & 0.65\\ 
				&    $t_y$    &    2.20    &    1.44    & 0.74      &        0.94       & \textbf{0.69} & 2.85 & \textbf{0.69} \\
				&    $t_z$    &    1.91    &    1.17    & 0.63        &    0.18   &   \textbf{0.17}   & 48.61 & 6.49  \\
				&\emph{$\alpha$}    &    85.81    &    1.32 &        1.28        &       0.35   &\textbf{ 0.12}   & 8.46 &1.5  \\
				&\emph{$\beta$}    &    42.12    &    0.75 &        1.08        &   0.24  &  \textbf{0.20} & 5.95  & 0.68 \\
				&\emph{$\gamma$}&    46.37    &    1.39    &     1.20        &       0.37    &\textbf{0.19} & 2.24  & 0.39 \\
				\midrule
				\multirow{6}{*}{{\rotatebox[origin=c]{90}{(d)~\emph{Tide}}}}
				&    $t_x$    &    1.46    &    0.853    &     0.69      &        0.73        &  \textbf{0.34 } & 1.31 & 1.74  \\
				&    $t_y$    &    2.25    &    1.37    & 0.81        &    0.56        &   \textbf{0.49 } &  0.83 & 0.74 \\ 
				&    $t_z$    &    0.92    &    1.20    & 0.81        &    0.24    &     \textbf{0.18} & 12.49 &   10.71 \\
				&\emph{$\alpha$}    &    5.15    &    1.78    &     2.10        &    0.31     &  \textbf{0.15 } & 2.03 &   1.78 \\
				&\emph{$\beta$}    &    2.13    &    1.09 &        1.38        &    \textbf{0.25    } &  0.39 &  1.56 & 1.64 \\
				&\emph{$\gamma$}&    2.98    &    1.13    &     1.27        &    \textbf{0.34    }   & 0.37 & 1.39 & 0.80	
			\end{tabular} 
		}
		& 
		\includegraphics[width=5cm]{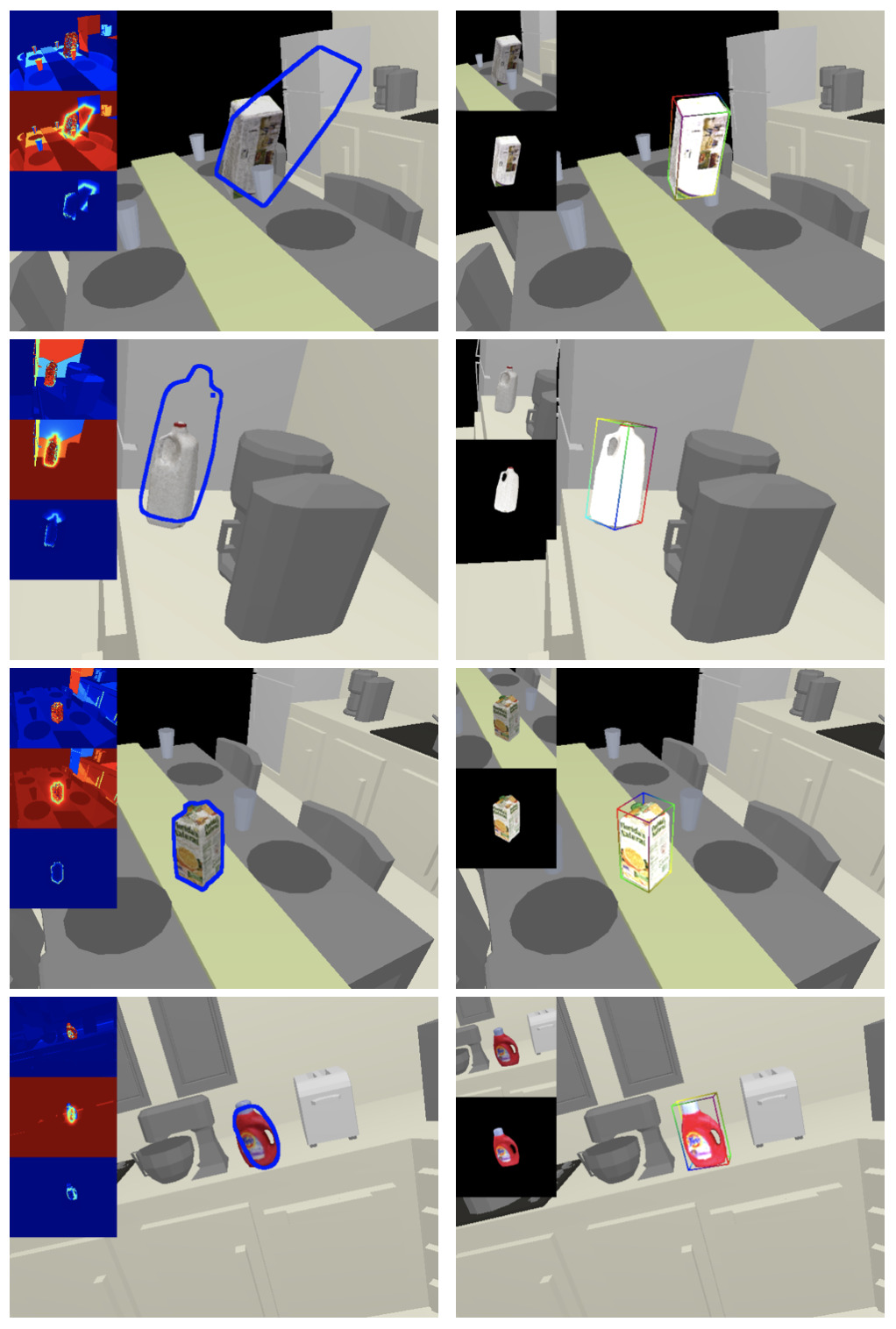} \\
		(a) Errors on 'Choi' in respect to others. & (b) Tracking quality compared to \cite{Tjaden2016}.
	\end{tabular}
	\caption{
		Left: Translation (mm) and rotation (degrees) errors on Choi for PCL's ICP, Choi and Christensen (C\&C)\cite{Choi2013}, Krull\cite{Krull2014}, Tan\cite{Tan2015}, Kehl\cite{Kehl2017a}, Tjaden\cite{Tjaden2016}  and our method. Right: Comparing \cite{Tjaden2016} (left) to us (right) using only RGB.
	}
	\label{tab:choi}
\end{figure} 

As a first use case we evaluated our method as a tracker on the 'Choi' benchmark \cite{Choi2013}. This RGB-D dataset consists of four synthetic sequences and we present detailed numbers in Figure \ref{tab:choi}. Note that all other methods utilize depth information. We decided for this dataset because it is very hard for RGB-only methods: it is poor in terms of color and the objects are of (semi-)symmetric nature. To provide an interesting comparison we also qualitatively evaluated against our tracker implementation of \cite{Tjaden2016}. While their method is usually robust for texture-less objects it diverges on 3 sequences which we show and for which we provide reasoning\footnote{The authors acknowledged our conclusions in correspondence.} in Figure \ref{tab:choi} and in the supplementary material. In essence, except for the 'Milk' sequence we can report very good results. The reason why we performed comparably bad on the 'Milk' resides in the fact that our method already treats it as a rather symmetric object. Thus, sometimes it rotates the object along its Y-axis, which has a negative impact on the overall numbers. In particular, while already being misaligned, the method still tries to completely fill the object into the scene, thus, it slightly further rotates and translates the object. 
Referring to the remaining objects, we can easily outperform PCL's ICP for all objects and also Choi and Christensen \cite{Choi2013} for most of the cases. Compared to Krull \cite{Krull2014}, which is a learned RGB-D approach, we perform better for some values and worse for others. Note that our translation error along the Z-axis is quite high. Since the difference in pixels is almost nonexistent when the object is moved only a few millimeters, it is almost impossible to estimate the exact distance of the object without leveraging depth information. This has also been discussed in \cite{Holloway:1997:REA:2871075.2871080} and is especially true for CNNs due to pooling operations.

\subsection{Detection refinement}

\begin{table}[t!]
	\centering
		\scalebox{0.82}{
			\begin{tabular}{@{}c|c|c|c|c|c|c|c|c|c|c|c|c|c||c@{}}
				& ape & bvise   & cam & can & cat  & driller & duck & box & glue & holep & iron & lamp & phone & total \\ \toprule
				No Refinement & 0.64 & 0.65 &  0.71 & 0.72 & 0.63 & 0.62 & 0.65 & 0.64 & 0.64 & 0.69 & 0.71 & 0.63 & 0.69 & 0.66 \\
				2D Edge-based ICP & 0.73 & 0.67	&  0.73 & 0.76 & 0.68  & 0.67	& 0.72 & 0.73 & 0.72 & 0.71 & 0.74 & 0.67 & 0.70 & 0.71 \\
				3D Cloud-based ICP & \textbf{0.86} & \textbf{0.88} & \textbf{0.91} & \textbf{0.87} & \textbf{0.87} & \textbf{0.85} & 0.83 & 0.84 & 0.75 & 0.77 & \textbf{0.85} & \textbf{0.84} & \textbf{0.81} & \textbf{0.84} \\
				Ours & 0.83 & 0.83 & 0.75 & \textbf{0.87} & 0.79 & \textbf{0.85} & \textbf{0.87} & \textbf{0.88} & \textbf{0.85} & \textbf{0.82} & \textbf{0.85} & 0.80 & 0.83 & 0.83 \\
			\end{tabular}
		}
		\caption{VSS scores for each sequence of \cite{Hinterstoisser2012} with poses initialized from SSD-6D \cite{Kehl2017}. The first three rows are provided by \cite{Kehl2017}. We evidently outperform 2D-based ICP by a large margin and are on par with 3D-based ICP.}
		\label{table:linemod_pose}
\end{table}

\begin{table}[t!]	
	\begin{tabular}{lr}
		\scalebox{0.81}{
			\begin{tabular}{@{}c|c|c|c}
				& Rot. Error [\degree] & Transl. Error [mm]   & ADD [\%] \\  \toprule
				No Ref. & 27.96 & 9.75, 9.33, 71.09 &  7.4 \\
				3D  ICP & 17.62 & 10.42, 10.56, \textbf{27.31} & \textbf{90.9}  \\
				Ours & \textbf{16.17} & \textbf{4.9}, \textbf{5.87}, 42.69 & 34.1  \\
				\cite{Rad2017} & -- & -- & 43.6 \\
				\cite{Brachmann2016} & -- & -- & 50.2 
				\\
				\toprule
				& Rot. Error [\degree] & Transl. Error [mm]   & ADD [\%]\\ \toprule
				No Ref. & 34.42 & 13.7, 13.4, 77.5 & 6.2 \\
				Ours & 24.36 & 8.5, 9.0, 49.1 & 27.5 \\
				
			\end{tabular}
		}
		 &
		\scalebox{0.81}{
			\begin{tabular}{c|c|c|c|c}
				Sequence & Ours & MSE Loss & Kehl \cite{Kehl2017a} & Tjaden \cite{Tjaden2016} \\
				\toprule 
				Camera 	& \textbf{0.803} &  0.562 & 0.493 & 0.385 \\
				Coffee & \textbf{0.848} & 0.717 & 0.747 & 0.170 \\
				Joystick & \textbf{0.850} & 0.746 & 0.773 & 0.298 \\
				Juice & \textbf{0.828} & 0.613 & 0.523 & 0.205 \\
				Milk & \textbf{0.766} & 0.721 & 0.580 & 0.514 \\ 
				Shampoo & \textbf{0.804} & 0.700 & 0.648 & 0.250 \\ \bottomrule 
				Total & \textbf{0.817} & 0.676 & 0.627 & 0.304 \\
			\end{tabular}
		} \\
		(a) Absolute pose errors on \cite{Hinterstoisser2012} and \cite{Brachmann2014}. & (b) VSS scores for each sequence of \cite{Tejani2014}.
	\end{tabular}
	
	\caption{Refinement scores with poses initialized from SSD-6D \cite{Kehl2017}. Left: Average ADD scores on 'Hinterstoisser' \cite{Hinterstoisser2012} (top) and 'Occlusion' \cite{Brachmann2014} (bottom).
		Right: VSS scores on 'Tejani'. We compare our visual loss to naive pose regression as well as two state-of-the-art trackers for the case of RGB \cite{Tjaden2016} and RGB-D \cite{Kehl2017a}.}
	\label{table:linemod_and_tejani_pose}
\end{table}

This set of experiments analyzes our performance in a detection scenario where an object detector will provide rough 6D poses and the goal is to refine them. We decided to use the results from SSD-6D \cite{Kehl2017}, an RGB-based detection method, that outputs 2D detections with a pool of 6D pose estimates each. The authors publicly provide their trained networks and we use them to detect and create 6D pose estimates which we feed into our system. Tables \ref{table:linemod_pose}, \ref{table:linemod_and_tejani_pose} (a) and (b) depict our results for the 'Hinterstoisser', 'Occlusion' and the 'Tejani' dataset using different metrics. We maximally ran 5 iterations of our method, yet, we also stopped if the last update was less than 1.5\degree and 7.5mm. Since our method is particularly strong at recovering from bad initializations, we employ the same RGB-verification strategy as SSD-6D. However, we apply it before conducting the refinement, since in contrast to them, we can also deal with imperfect initializations, as long as they are not completely misaligned.
We report our errors with the VSS metric (which is VSD from \cite{Hodan2016} with $\tau=\infty$) that calculates a visual 2D error as the pixel-wise overlap between the renderings of ground truth pose and estimated pose. Furthermore, to compare better to related work, we also use the ADD score  \cite{Hinterstoisser2012} to measure a 3D metrical error as the average point cloud deviation between real pose and inferred pose when transformed into the scene. A pose is counted as correct if the deviation is less than a $\frac{1}{10}$th of the object diameter. 

Referring to 'Hinterstoisser' with the VSS metric, we can strongly improve the state-of-the-art for most objects. In particular, for the case of RGB only, we can report an average VSS score of 83\%, which is an improvement of impressive and can thus successfully bridge the gap between RGB and RGB-D in terms of pose accuracy.

Except for the 'cam' and the 'cat' object our results are on par with or even better than SSD-6D + 3D refinement. ICP relies on good correspondences and robust outlier removal which in turn requires very careful parameter tuning. Furthermore, ICP is often unstable for rougher initializations. In contrast, our method learns refinement end-to-end and is more robust since it adapts to the specific properties of the object during training. However, due to this, our method requires meshes of good quality. Hence, similar to SSD-6D we have especially problems for the 'cam' object since the model appearance strongly differs from the real images which exacerbates training. Also note that their 3D refinement strategy uses ICP for each pose in the pool, followed by a verification over depth normals to decide for the best pose. Our method instead uses a simple check over image gradients to pick the best. 

\begin{figure}[t!]
	\centering
	\includegraphics[width=2.95cm]{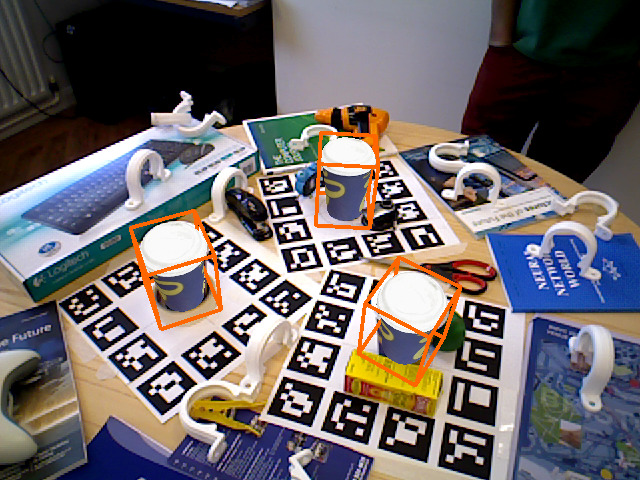}
	\includegraphics[width=2.95cm]{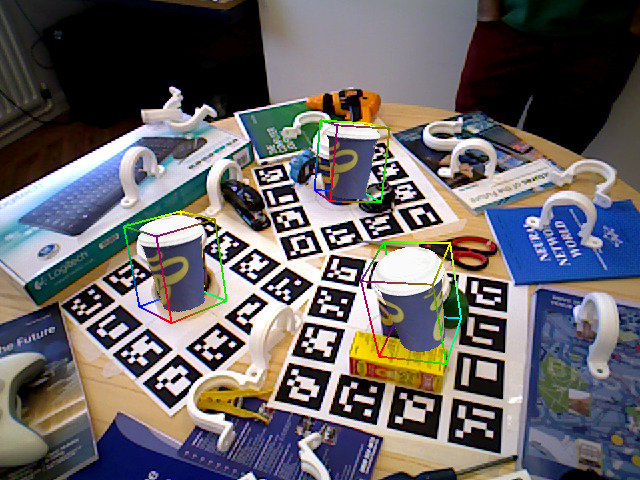}
	\includegraphics[width=2.95cm]{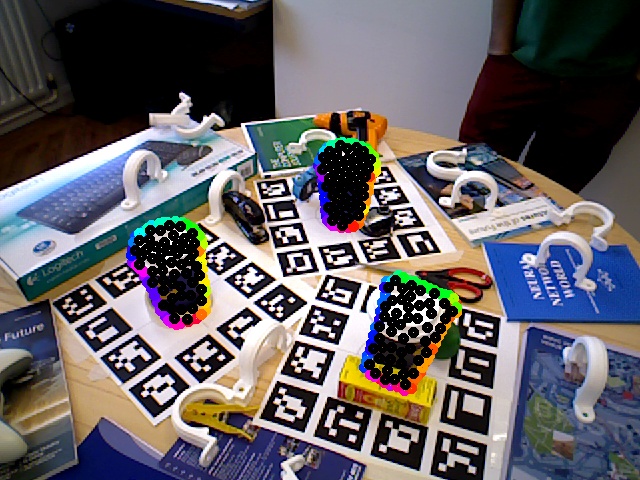}
	\includegraphics[width=2.95cm]{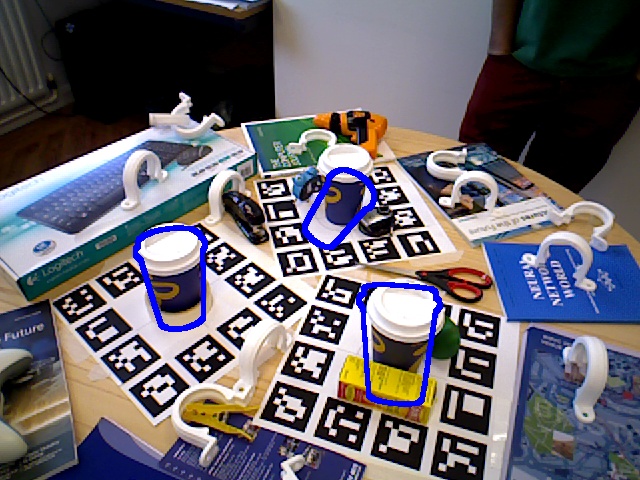}		
	
	\caption{Comparison on Tejani between (from left to right) our visual loss, mean squared error loss, the RGB-D tracker from \cite{Kehl2017a} and the RGB tracker from \cite{Tjaden2016}.}
	\label{fig:tejani}
\end{figure}

With respect to the ADD metric we fall slightly behind the other state-of-the-art RGB methods \cite{Brachmann2016,Rad2017}. We got the 3D-ICP refined poses from the SSD-6D authors and analyzed the errors in more detail in Table \ref{table:linemod_and_tejani_pose}(a). We see again that we have bigger errors along the Z-axis, but less errors along X and Y. Unfortunately, the ADD metric penalizes this deviation overly strong. Interestingly, \cite{Brachmann2016,Rad2017} have better scores and we reason this to come from two facts. The datasets are annotated via ICP with 3D models against depth data. Unfortunately, inaccurate intrinsics and the sensor registration error between RGB and D leads to an inherent mismatch where the ICP 6D pose does not always align perfectly in RGB. Purely synthetic RGB methods like ours or \cite{Kehl2017} suffer from (1) a domain gap in terms of texture/shape and (2) the dilemma that better RGB performance can worsen results when comparing to that 'true' ICP pose. We suspect that \cite{Brachmann2016,Rad2017} can learn this registration error implicitly since they train on real RGB cut-outs with associated ICP pose information and thus avoid both problems. We often observe that our visually-perfect alignments in RGB fail the ADD criterion and we show examples in the supplement. 
Since our loss actually optimizes a form of VSS to maximize contour overlap, we can expect the ADD scores to go up only when perfect alignment in color equates perfect alignment in depth.

Eventually, referring to the 'Occlusion' dataset, we can report a strong improvement compared to the original numbers from SSD-6D, despite the presence of strong occlusion. In particular, while the rotational error decreased by approximately 8 degrees, the translational error dropped by 4mm along 'X' and 'Y' axes and by 28mm along 'Z'. Thus, we can increase ADD from 6.2\% up to 28.5\%, which demonstrates that we can deal with strong occlusion in the scene.

For 'Tejani' we decided to show the improvement over networks trained with a standard regression loss (MSE). Additionally, we re-implemented the RGB tracker from \cite{Tjaden2016} and were kindly provided with numbers from the authors of the RGB-D tracker from \cite{Kehl2017a} (see Figure \ref{fig:tejani}). Since the dataset mostly consists of objects with geometric symmetry, we do not measure absolute pose errors here but instead report our numbers with the VSS metric. The MSE-trained networks constantly underperform since the dataset models are of symmetric nature which in turn leads to a large difference of 14\% in comparison to our visual loss. This result stresses the importance of correct symmetry entangling during training. The RGB tracker was not able to refine well due to the fact that the color segmentation was corrupted by either occlusions or imperfect initialization. The RGB-D tracker, which builds on the same idea, performed better because it uses the additional depth channel for segmentation and optimization.

\subsection{Category-level tracking}

\begin{figure}[t]
	\begin{center}
		\includegraphics[width=12cm]{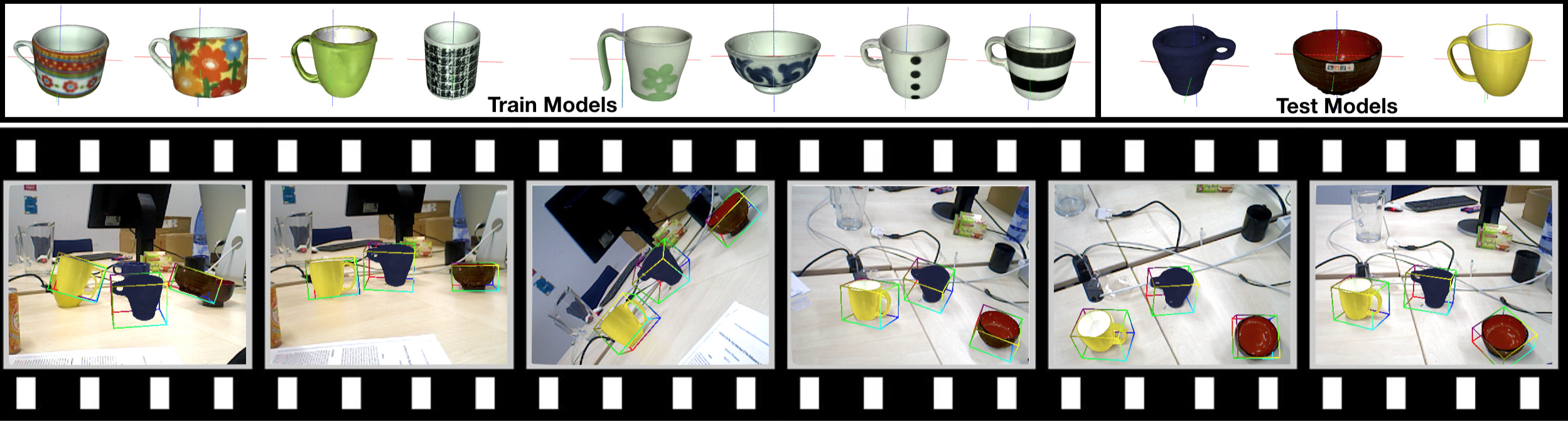}
	\end{center}
	\caption{Qualitative category-level experiment where we train our network on a specific set of mugs and bowls and track hitherto unseen models. The first frame depicts very rough initialization while the next frames show some intermediate refined poses throughout the sequence. The supplement shows the full video.}
	\label{fig:class}
\end{figure}
We were curious to find out whether our approach can generalize beyond a specific CAD model, given that many objects from the same category share similar appearance and shape properties. To this end, we conducted a final qualitative experiment (see Figure \ref{fig:class}) where we collected a total of eight CAD models of cups, mugs and a bowl and trained simultaneously on all. During testing we then used this network to track new, unseen models from the same category. We were surprised to see that the approach has indeed learned to metrically track previously unseen but nonetheless similar structures. While the poses are not as accurate as for the single-instance case, it seems that one can indeed learn the projective relation of structure and how it changes under 6D motion, provided that at least the projection functions (i.e. camera intrinsics) are constant. We show the full sequence in the supplementary material.

\subsection{Failure cases}
\begin{figure}[t!]
	\begin{center}	
		\includegraphics[width=2.95cm]{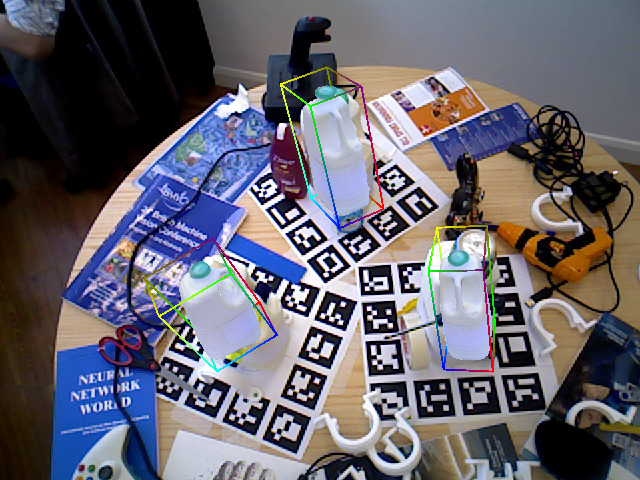}
		\includegraphics[width=2.95cm]{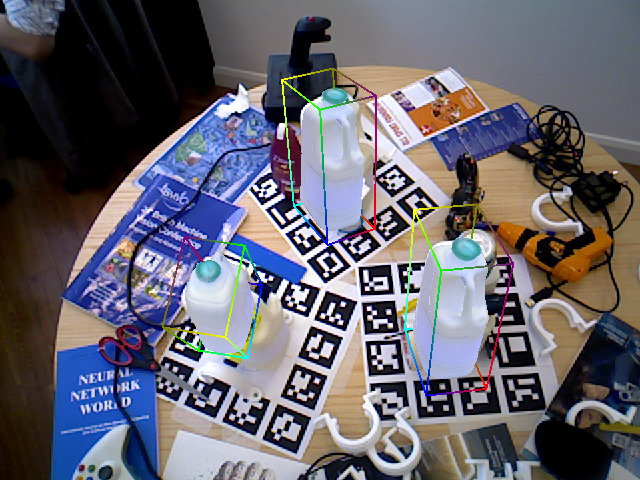}
		\includegraphics[width=2.95cm]{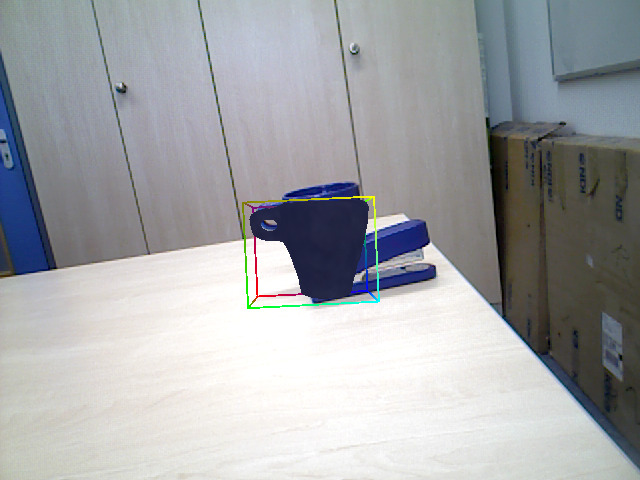}
		\includegraphics[width=2.95cm]{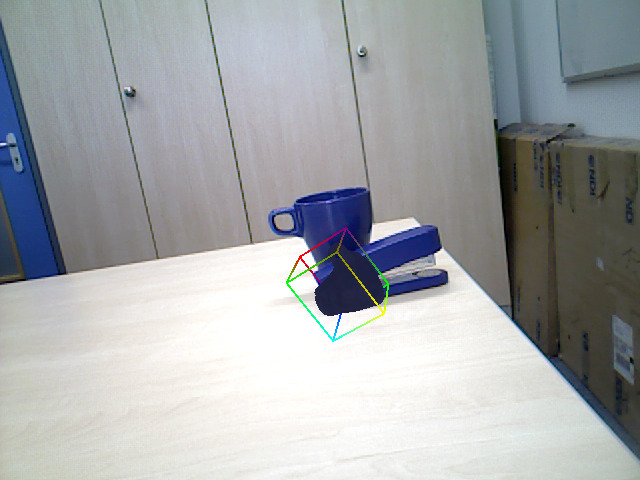}
	\end{center}
	\caption{Two prominent failure cases: Occlusion (left pair) and objects of very similar colors and shapes (right pair) can negatively influence the regression.}
	\label{fig:failure}
\end{figure}

Figure \ref{fig:failure} illustrates two known failure cases where the left image of each pair represents initialization and the right image the refined result. Although we train with occlusion certain occurrences can worsen our refinement nonetheless. While two 'milk' instances were refined well despite occlusion, the left 'milk' instance could not be recovered correctly. The network assumes the object to end at the yellow pen and only maximizes the remaining pixel-wise overlap. Besides occlusion, objects of similar color and shape can in rare cases lead to confusion. As shown in the right pair, the network mistakenly assumed the stapler, instead of the cup, to be the real object of interest.

%% file: main.bbl
\begin{thebibliography}{10}
\providecommand{\url}[1]{\texttt{#1}}
\providecommand{\urlprefix}{URL }
\providecommand{\doi}[1]{https://doi.org/#1}

\bibitem{Abadi2016}
Abadi, M., Barham, P., Chen, J., Chen, Z., Davis, A., Dean, J., Devin, M.,
  Ghemawat, S., Irving, G., Isard, M., Kudlur, M., Levenberg, J., Monga, R.,
  Moore, S., Murray, D., Steiner, B., Tucker, P., Vasudevan, V., Warden, P.,
  Wicke, M., Yu, Y., Zheng, X.: {TensorFlow: Large-scale machine learning on
  heterogeneous systems}. In: OSDI (2016),
  \url{http://download.tensorflow.org/paper/whitepaper2015.pdf}

\bibitem{Bhagavatula2017}
Bhagavatula, C., Zhu, C., Luu, K., Savvides, M.: {Faster Than Real-time Facial
  Alignment: A 3D Spatial Transformer Network Approach in Unconstrained Poses}.
  In: ICCV (2017), \url{http://arxiv.org/abs/1707.05653}

\bibitem{Bibby2008}
Bibby, C., Reid, I.: {Robust Real-Time Visual Tracking using Pixel-Wise
  Posteriors}. In: ECCV (2008)

\bibitem{Brachmann2014}
Brachmann, E., Krull, A., Michel, F., Gumhold, S., Shotton, J., Rother, C.:
  {Learning 6D Object Pose Estimation using 3D Object Coordinates}. In: ECCV
  (2014)

\bibitem{Brachmann2016}
Brachmann, E., Michel, F., Krull, A., Yang, M.Y., Gumhold, S., Rother, C.:
  {Uncertainty-Driven 6D Pose Estimation of Objects and Scenes from a Single
  RGB Image}. In: CVPR (2016)

\bibitem{Brox2010}
Brox, T., Rosenhahn, B., Gall, J., Cremers, D.: {Combined Region and
  Motion-Based 3D Tracking of Rigid and Articulated Objects}. TPAMI  (2010)

\bibitem{Choi2013}
Choi, C., Christensen, H.: {RGB-D Object Tracking: A Particle Filter Approach
  on GPU}. In: IROS (2013)

\bibitem{Dambreville2010}
Dambreville, S., Sandhu, R., Yezzi, A., Tannenbaum, A.: {A Geometric Approach
  to Joint 2D Region-Based Segmentation and 3D Pose Estimation Using a 3D Shape
  Prior}. SIAM Journal on Imaging Sciences  (2010)

\bibitem{Drummond2002}
Drummond, T., Cipolla, R.: {Real-time visual tracking of complex structures}.
  TPAMI  (2002)

\bibitem{Garon2017}
Garon, M., Lalonde, J.F.: {Deep 6-DOF Tracking}. In: ISMAR (2017).
  \doi{10.1109/TVCG.2017.2734599}

\bibitem{Hexner2016}
Hexner, J., Hagege, R.R.: {2D-3D Pose Estimation of Heterogeneous Objects Using
  a Region Based Approach}. IJCV  (2016)

\bibitem{Hinterstoisser2012}
Hinterstoisser, S., Lepetit, V., Ilic, S., Holzer, S., Bradski, G., Konolige,
  K., Navab, N.: {Model based training, detection and pose estimation of
  texture-less 3D objects in heavily cluttered scenes}. In: ACCV (2012)

\bibitem{Hinterstoisser2017}
Hinterstoisser, S., Lepetit, V., Wohlhart, P., Konolige, K.: On pre-trained
  image features and synthetic images for deep learning. CoRR
  \textbf{abs/1710.10710} (2017), \url{http://arxiv.org/abs/1710.10710}

\bibitem{Hodan2016}
Hodan, T., Matas, J., Obdrzalek, S.: {On Evaluation of 6D Object Pose
  Estimation}. In: ECCV Workshop (2016)

\bibitem{Holloway:1997:REA:2871075.2871080}
Holloway, R.L.: Registration error analysis for augmented reality. Presence:
  Teleoper. Virtual Environ.  \textbf{6}(4),  413--432 (Aug 1997).
  \doi{10.1162/pres.1997.6.4.413},
  \url{http://dx.doi.org/10.1162/pres.1997.6.4.413}

\bibitem{Jaderberg2015}
Jaderberg, M., Simonyan, K., Zisserman, A., Kavukcuoglu, K.: {Spatial
  Transformer Networks}. In: NIPS (2015), \url{http://arxiv.org/abs/1509.05329}

\bibitem{Kehl2017}
Kehl, W., Manhardt, F., Ilic, S., Tombari, F., Navab, N.: {SSD-6D: Making
  RGB-Based 3D Detection and 6D Pose Estimation Great Again}. In: ICCV (2017)

\bibitem{Kehl2017a}
Kehl, W., Tombari, F., Ilic, S., Navab, N.: {Real-Time 3D Model Tracking in
  Color and Depth on a Single CPU Core}. In: CVPR (2017)

\bibitem{Kendall2017}
Kendall, A., Cipolla, R.: {Geometric loss functions for camera pose regression
  with deep learning}. In: CVPR (2017), \url{http://arxiv.org/abs/1704.00390}

\bibitem{Kendall2015}
Kendall, A., Grimes, M., Cipolla, R.: {PoseNet: A Convolutional Network for
  Real-Time 6-DOF Camera Relocalization}. In: ICCV (2015)

\bibitem{Krull2014}
Krull, A., Michel, F., Brachmann, E., Gumhold, S., Ihrke, S., Rother, C.:
  {6-DOF Model Based Tracking via Object Coordinate Regression}. In: ACCV
  (2014)

\bibitem{Liu2016}
Liu, W., Anguelov, D., Erhan, D., Szegedy, C., Reed, S., Fu, C.y., Berg, A.C.:
  {SSD : Single Shot MultiBox Detector}. In: ECCV (2016)

\bibitem{Park2008}
Park, Y., Lepetit, V.: Multiple 3d object tracking for augmented reality. In:
  ISMAR (2008)

\bibitem{Pauwels2013}
Pauwels, K., Rubio, L., Diaz, J., Ros, E.: {Real-time model-based rigid object
  pose estimation and tracking combining dense and sparse visual cues}. In:
  CVPR (2013)

\bibitem{Pavlakos2017}
Pavlakos, G., Zhou, X., Chan, A., Derpanis, K.G., Daniilidis, K.: {6-DoF Object
  Pose from Semantic Keypoints}. In: ICRA (2017),
  \url{http://arxiv.org/abs/1703.04670}

\bibitem{Prisacariu2015}
Prisacariu, V.A., Murray, D.W., Reid, I.D.: {Real-Time 3D Tracking and
  Reconstruction on Mobile Phones}. TVCG  (2015)

\bibitem{Prisacariu2012}
Prisacariu, V.A., Reid, I.D.: {PWP3D: Real-Time Segmentation and Tracking of 3D
  Objects}. IJCV  (2012)

\bibitem{Rad2017}
Rad, M., Lepetit, V.: {BB8:} {A} scalable, accurate, robust to partial
  occlusion method for predicting the 3d poses of challenging objects without
  using depth. In: ICCV. pp. 3848--3856 (2017). \doi{10.1109/ICCV.2017.413},
  \url{https://doi.org/10.1109/ICCV.2017.413}

\bibitem{Rosenhahn2006}
Rosenhahn, B., Brox, T., Cremers, D., Seidel, H.P.: {A comparison of shape
  matching methods for contour based pose estimation}. LNCS  (2006)

\bibitem{Schmaltz2007}
Schmaltz, C., Rosenhahn, B., Brox, T., Cremers, D., Weickert, J., Wietzke, L.,
  Sommer, G.: {Region-Based Pose Tracking}. In: IbPRIA (2007)

\bibitem{Schmaltz2012}
Schmaltz, C., Rosenhahn, B., Brox, T., Weickert, J.: {Region-based pose
  tracking with occlusions using 3D models}. MVA  (2012)

\bibitem{Seo2014}
Seo, B.K., Park, H., Park, J.I., Hinterstoisser, S., Ilic, S.: {Optimal local
  searching for fast and robust textureless 3D object tracking in highly
  cluttered backgrounds}. In: TVCG (2014)

\bibitem{Szegedy2016}
Szegedy, C., Ioffe, S., Vanhoucke, V., Alemi, A.A.: Inception-v4,
  inception-resnet and the impact of residual connections on learning. In: ICLR
  Workshop (2016), \url{https://arxiv.org/abs/1602.07261}

\bibitem{Tan2015}
Tan, D.J., Tombari, F., Ilic, S., Navab, N.: {A Versatile Learning-based 3D
  Temporal Tracker : Scalable , Robust , Online}. In: ICCV (2015)

\bibitem{Tateno2009}
Tateno, K., Kotake, D., Uchiyama, S.: {Model-based 3D Object Tracking with
  Online Texture Update}. In: MVA (2009)

\bibitem{Tejani2014}
Tejani, A., Tang, D., Kouskouridas, R., Kim, T.k.: {Latent-class hough forests
  for 3D object detection and pose estimation}. In: ECCV (2014)

\bibitem{Tjaden2016}
Tjaden, H., Schwanecke, U., Schoemer, E.: {Real-Time Monocular Segmentation and
  Pose Tracking of Multiple Objects}. In: ECCV (2016)

\bibitem{Tjaden2017}
Tjaden, H., Schwanecke, U., Sch{\"{o}}mer, E.: {Real-Time Monocular Pose
  Estimation of 3D Objects using Temporally Consistent Local Color Histograms}.
  In: ICCV (2017). \doi{10.1109/ICCV.2017.23}

\bibitem{Ummenhofer2017}
Ummenhofer, B., Zhou, H., Uhrig, J., Mayer, N., Ilg, E., Dosovitskiy, A., Brox,
  T.: {DeMoN: Depth and Motion Network for Learning Monocular Stereo}. In: CVPR
  (2017)

\bibitem{Vacchetti2004}
Vacchetti, L., Lepetit, V., Fua, P.: {Stable Real-Time 3D Tracking Using Online
  and Offline Information}. TPAMI  (2004)

\bibitem{Wang2017}
Wang, S., Clark, R., Wen, H., Trigoni, N.: {DeepVO: Towards End to End Visual
  Odometry with Deep Recurrent Convolutional Neural Networks}. In: ICRA (2017)

\bibitem{Wu2016}
Wu, J., Xue, T., Lim, J.J., Tian, Y., Tenenbaum, J.B., Torralba, A., Freeman,
  W.T.: {Single Image 3D Interpreter Network.} In: ECCV (2016).
  \doi{10.1007/978-3-319-46466-4},
  \url{http://dblp.uni-trier.de/db/conf/eccv/eccv2016-6.html{\#}0001XLTTTF16}

\bibitem{Zhou2017}
Zhou, T., Brown, M., Snavely, N., Lowe, D.G.: {Unsupervised Learning of Depth
  and Ego-Motion from Video}. In: CVPR (2017),
  \url{http://arxiv.org/abs/1704.07813}

\end{thebibliography}
